# A syllable based model for handwriting recognition

*Wassim Swaileh*
Normandie University, University of Rouen
LITIS Laboratory EA 4108
Rouen, France
Wassim.swaileh2@univ-rouen.fr

*Thierry Paquet*
Normandie University, University of Rouen
LITIS Laboratory EA 4108
Rouen, France
Thierry.Paquet@univ-rouen.fr

*Abstract*— In this paper, we introduce a new modeling approach of texts for handwriting recognition based on syllables. We propose a supervised syllabification approach for the French and English languages for building a vocabulary of syllables. Statistical n-gram language models of syllables are trained on French and English Wikipedia corpora. The handwriting recognition system, based on optical HMM context independent character models, performs a two pass decoding, integrating the proposed syllabic models. Evaluation is carried out on the French RIMES dataset and English IAM dataset by analyzing the performance for various coverage of the syllable models. We also compare the syllable models with lexicon and character n-gram models. The proposed approach reaches interesting performances thanks to its capacity to cover a large amount of out of vocabulary words working with a limited amount of syllables combined with statistical n-gram of reasonable order.

*Keywords—syllable; syllabification; handwriting; language model; recognition*

## I. Introduction

Handwriting recognition is investigated for several decades in order to transform images of handwritten texts into their digital transcriptions encoded in ASCII or Unicode. In this purpose the general idea of handwriting recognition systems is to represent the image properties by probabilistic models that are optical models of the characters of the concerned language to be recognized. Through a learning process, the optical models are optimized on a set of annotated examples in order to achieve the best transcription. In the literature, various structures have been proposed to perform handwriting recognition [21]. According to their lexical structure, the recognition systems can be classified into three main categories.

The first category includes the closed vocabulary systems that are optimized for the recognition of words in a limited and static vocabulary. This kind of system is used for specific applications such as bank checks reading [7]. In this case the optical models may be word models. These approaches are often mentioned as global or holistic recognition.

The second category includes dynamic vocabulary systems that are able to recognize words never seen by the system during training. In this category, the optical models are character models, and the approach is often referred as analytical recognition approaches that are guided by the knowledge of a lexicon (lexicon driven) during the recognition phase. With this capability, these systems are used for general purpose applications, such as recognition of historical documents, for example [13]. As a result of the increase of the vocabulary size, the recognition task becomes more difficult, and its performance decreases gradually due to the introduction of more and more resembling words that must be discriminated by the system [10]. To compensate this decrease in performance, it is possible to constrain the recognition system by a model of phrases that models word sequences. This is known as language model, and the complexity of the recognition system increases further, since it has to manage simultaneously the optical models, the vocabulary, and the language model [21].

The third category of approaches includes systems without vocabulary (lexicon free) that perform recognition in lines of text by recognizing sequences of characters. To improve their performance, these systems may use character sequences models in the form of n-gram statistical models [21] by considering the space between words as a character [4]. The advantage of these systems is their ability to recognize any sequence of characters, including out of vocabulary words (OOV) such as named entities. Hence, they have the disadvantage of being less efficient than previous models in the absence of the sentence level modeling. Kozielski & al. [11] have explored the use of character language models (for English and Arabic) using 10-gram character models estimated using the Witten-Bell method. They compared this lexicon free approach with a lexicon based approach associated with a 3-gram language model estimated using the modified Kneser-Ney estimation method. They have also combined the two models (characters and words) by using two approaches. The first one by building a global interpolated model of the two models, the second one by using a combination of back-off models. The results show the effectiveness of the combination of the two language models using interpolation.

Generally, in order to reduce the negative effect of OOV words, one try to increase the vocabulary size which controls the recognition system, but this is at the expense of computational complexity and confusion [16], having in mind that the vocabulary is never complete anyway. An alternative approach in order to optimize the performance / lexicon size tradeoff, may be to use sub-word units models. In this case, recognition of OOV becomes possible due to the lexicon coverage ability of sub-words units [14]. In addition, the system may be guided by a specific language model operating on sub-words units. This approach is interesting only for sufficiently inflected languages, such as Arabic language and several approaches have been proposed in the literature. Hamdani & al. [9] proposed an Arabic handwriting recognition system based on HMM models including a part of Arabic words language model. The vocabulary contains words and sub-word parts produced by a specific morphological decomposition method of Arabic words. The decomposition is based on the morphological definition of roots, suffixes and prefixes of words [6]. The results show the improvement provided by this system, notably to cope with OOV, compared to a lexicon driven recognition system.

BenZeghiba & al. [3] proposed a hybrid model for Arabic language, which is designed according to the observed frequency of words. The idea is to keep the most frequent words as is (without decomposition), and to decompose only the least frequent words into sub-words of PAWs. By taking advantage of the specific property of the Arabic language and script, a PAW (Part of Arabic Word) is a sequence of characters that can be connected together. A character that cannot be connected with its following neighbor defines the end of a PAW. The advantage of a hybrid word - PAW model lies in the tradeoff between its modeling ability and its reduced complexity. The two models (hybrid word + PAW) and PAW alone perform almost similarly on OOV words but the hybrid system is less complex.

Jacqueline & al. [5] proposed a probabilistic syllable model for scene text recognition in different image conditions such as advertisements panels. The proposed model was kept away from using n-grams language models. They used a probabilistic contextual free grammar (PCFG) that models each syllable as a sequence of consonant and vowel groups of characters. The used PCFG encapsulate information about syllables (English consonant and vowel groups) which forces a word label to be consistent with a grammar. The proposed model shows a good performance for the OOV words such as proper nouns, under the condition that these words should all be pronounceable in English. By the definition of special parsing roles of characters and training process using a syllabified dictionary the proposed PCFG model was built. As a defect of this model, the grammar does not use any contextual information about syllables, such as n-gram model. This model was tested for isolated words recognition on two different data sets and its performance were compared with the performance of lexicon directed recognition system that use a bi-gram character language model. The proposed model shows an improvement of 4% compared to the lexicon directed recognition system accuracy.

Inspired by these early works, and some others devoted to speech recognition, we propose in this communication a modeling approach of handwriting based on a model of sub-word units. This model is based on an orthographic syllable model of the written language which itself relies on phonetic modeling. The challenge is to produce a language syllable model of reasonable size that is able to constrain the optical recognition system efficiently in order to achieve good performance on OOV words. The first step of our approach lies in the definition of a lexicon of syllables. In this purpose we propose a supervised automatic orthographic syllabification method that exploit some computerized resources in order to generate a list a syllables from an input lexicon of words. Then, the second step of our approach computes statistical n-gram syllable language models from a training corpus. Various language models can be estimated for different word coverage rates of the test set, thus allowing evaluating the performance of the recognition system for various OOV rates. Finally these language models are integrated into a HMM based handwriting recognition system for evaluation on two datasets. Experimentations are carried out for French and English using the RIMES 2011 dataset [8] and the IAM dataset [1] respectively.

The organization of this article is the following one: the theoretical basis of syllabic model is presented in part 2, the proposed syllabification method is described in part 3. We present the structure of the recognition system in part 4. The experiments are presented and analyzed in part 5, before drawing some perspectives of this work.

II. SYLLABLE BASED MODELLING APPROACH

Syllables play an important role in the organization of speech and language [2-23]. The name "syllable" is sometimes defined physiologically as a continuous unit of the spoken language, which consists of a sound or group of sounds uttered in one breath [5, 26]. The segmentation of speech into syllables can be achieved using acoustic units or phonological units [15], and syllables produced by these two models are not always compatible [23]. Most phoneticians agree that a syllable is composed basically of a rhyme that is preceded by an onset (one or more consonants "C" optionally comes at the beginning of the syllable). Inside a rhyme, the nucleus (usually a vowel "V") is the constitutive element of the syllable. This is followed by a coda (one or more consonants "C" at the end of the syllable) [23]. The languages differ from each other with respect to topological parameters as optionality of the onset and admissibility of the codas. For example, the onsets are mandatory in German while the codas are prohibited in Spanish [18]. In French and English, the nucleus is always considered as a vowel. Thus, counting the number of syllables pronounced in a French or English, should be equivalent to counting the number of pronounced vowels [23].

Considering written languages, and according to [19], the orthographic syllable differs from the phonetic syllable because it retains all "*e*" silent placed between two consonants or placed at the end of a word. Hyphenation rules separate the double consonants even if they are pronounced as a single consonant. For example, graphically there are three syllables in the French word ***pu-re-té*** even if we pronounce it as [ *pyr-te* ] (two phonetic syllables). The authors of [22] have classified the syllables in three different categories:

- A ***phonetic syllable*** is composed of a combination of phonemes that are pronounced in a single breath.
- A ***graphemic syllable*** represents a faithful transposition of phonetic syllabification in the spelling of the word.
- An ***orthographic syllable*** applies hyphenation rules that must be adhered to writing.

It seems difficult to conciliate these different views of specialists, but in any case, only graphemic or orthographic syllables provide a decomposition of writing that is likely to have an impact on a writing recognition system. In this study, we chose to use the orthographic representation of syllables provided by the French computerized and syllabified lexical database *Lexique3* [12], and the free English language hyphenation dictionary [29], which was adapted from Grady Ward's public domain English hyphenation dictionary [26]. The *Lexique3* database provides an orthographic syllabic decomposition of a lexicon of almost 142 695 French words into 9 522 syllables only. It therefore constitutes a knowledge base from which our French syllabic model is developed. The free English language hyphenation dictionary contains 166280 words decomposed into 21 991 syllables.

However, despite their relatively large size, it quickly becomes out that these databases by fare do not cover the French or English vocabularies. For example, the *Lexique3* database covers only 69.83% of the vocabulary of the RIMES database, which is one of the training and reference dataset for French handwriting recognition. Similarly, the free English hyphenation dictionary covers only 54.42% of the IAM database vocabulary.

Therefore, we have to find a general way for generating a syllabic decomposition of any corpus. For this purpose it is necessary to develop an automatic syllabification method. The method we propose is a supervised method[1] that exploits the Lexique3 database or the free English hyphenation dictionary and a measure of similarity between words.

## III. AN AUTOMATIC SUPERVISED SYLLABIFICATION METHOD

The method that we propose is based on matching similar lexical and phonetic structures in order to provide a syllabic decomposition of any unknown word. Let us assume a lexicon of words $L$, each word entry being represented by its sequence of characters $m_n$ associated with its syllabic decomposition denoted $s_n$, namely its sequence of syllables. Formally we can write $L = \{(m_1, s_1), (m_2, s_2), \ldots, (m_m, s_m), \ldots, (m_l, s_l), \}$.

For any unknown word (not in $L$) represented by its character string $m$, we wish to determine its syllable representation $s$. The first idea is to search for the closest word $m_n$ in the dictionary to the unknown input character string. But two very similar words may have different syllabic decomposition especially if they differ by a vowel, which often marks the presence of a syllable. To take this information into account, we introduce an orthographic structure that represents the word by its sequence in consonants and vowels. For example, the word "***Bonjour***" is encoded by its orthographic structure $ss$ = "***CVCCVVC***". Then we can define a similarity measure by combining both representations according to equation 1, where $S_{lex}$ and $S_{syl}$ are respectively two similarity measures on the lexical and the syllabic representations:

$$S_G\big((m, ss), (m_i, ss_i)\big) = \frac{S_{lex}(m, m_i) + S_{syl}(ss, ss_i)}{2} \quad (1)$$

The similarity measure between sequences of characters counts the average number of identical character pairs at the same positions between the two sequences. When the two strings are of different length, the shorter is completed at the end by empty characters, so the two sequences have an identical size. By this way, we make the division into syllables based on the prefix of the dictionary word, the errors of cutting words into syllables may occur on word suffixes that may be different due to of the completion. When the most similar lexicon entry to the unknown word gets a similarity score above a threshold $T$, its syllabic representation serves as a model to decompose the unknown word. More precisely, the syllables segmentation of the unknown word is made at the same positions as in the lexicon entry. When the similarity score is below the threshold, the decomposition into characters is admitted as syllabic decomposition by default. Table 1 gives some examples of words syllabification for some words that have no syllable decomposition in *Lexique3* nor in the English hyphenation dictionary.

|    | Query word | Database candidate and its syllabifications | Proposed syllabification |
|---|---|---|---|
| FR  | Bonjour | Toujours (Tou-jours) | Bon-jour |
|    | dérogatoire | dédicatoire (dé-di-ca-toi-re) | dé-ro-ga-toi-re |
| EN  | moved | moped (mo-ped) | mo-ved |
|    | Answering | anglewing (an-gle-wing) | An-swe-ring |

TABLE I. EXAMPLES OF SYLLABIFIED WORDS BY THE METHOD.

## IV. THE HANDWRITING RECOGNITION SYSTEM

Our recognition system is based on optical modeling of the characters based on hidden Markov models (HMM). The essential components in the design of our system are the alphanumeric characters. We have in total, 100 character models for the RIMES dataset and 80 character models for the IAM dataset, by considering the white space between words as a character. Our recognition system is composed of four main stages; namely, preprocessing, optical models training, lexicon and syllable lexicon generation and language model training. The recognition step is performed following a two passes decoding algorithm.

### A. Pre-procssing

During preprocessing we proceed to the detection of text lines in the text blocks in order to improve the rectangular positioning provided in the RIMES dataset because it provides quite noisy lines. Indeed rectangular areas provide lines that contain overlaps with the line above or the line below. The automatic method for line surrounding is described in [27]. Considering the IAM dataset, the rectangular zoning provides quite clean lines. Then line images are adjusted horizontally and vertically (*deskew* and *deslant*) and scaled to a 96 pixels height, preserving aspect ratio.

### B. Optical Character Models

Optical models exploit HoG (Histogram of Gradients) characteristics extracted from the text line images by using a sliding window of 20 pixels width (a frame). Frame horizontal displacement is 2 pixels. Each frame is described with a vector composed of 70 real features. 64 features encode the HoG description, and 6 encode a geometric description of the frame. Generally, the internal structure of the character optical Models (HMM) is defined by a fixed number of hidden states and for each of them, a fixed size Gaussian mixture is also determined.

---
[1] The source code of the proposed method could be forked from http://swaileh.github.io/Syllabifier

We chose to use mixtures of 20 Gaussians, which guarantee a description ability fairly accurate for each frame. Determining the number of hidden states is an optimization problem. An overestimated number of the hidden states leads to over-trained models. An underestimated number of states leads to inadequate specialized models. This problem has been addressed in [28], [25], [24]. We have been inspired by the proposed method in the first reference that is based on the Bakis method in order to optimize the number of states of each character model. Once a first training of one initial set of character models has been carried out with a fixed number of states, we compute the average number of frames $F$ per optical model using a forced alignment of the corresponding model on the ground truth of each image on the frame sequence. The number of states E of the corresponding model is then defined as a fraction of $F$ ($E = \alpha.F$).

A new training process (*Baum-Welch*) is performed for the new parameterized models that have been created, according to parameter $\alpha$. Then we perform a final decoding without ground truth (no forced alignment) using the trained models and the character recognition rate (CRR) is computed. The operation is repeated for different values of α (increasing values between 0 and 1). Finally we select the most accurate models based on a criterion combining the average CRR and the alignment rates of the models on the training examples. Indeed, excessively long character models tend to maximize the recognition rate but at the expense of misalignments on shorter examples. This criterion is tested at each iteration of the *Baum-Welch* training procedure. Training is stopped when the criterion reaches its highest value. We then obtain optimized optical models. Training the optical models is performed on the RIMES 2011 training dataset, which contains 10,963 images of tagged text lines that are segmented from 1500 images of paragraphs written by different writers in different writing conditions.

*C. lexicons and language models*

The third step in building our system is the definition of the vocabularies and language models to be used by the system during decoding. Two text corpora are used for vocabularies and language models generation. The initial corpuses are collected from the texts of the RIMES dataset (learning set and validation set) and the IAM dataset. A second corpora, much larger in size, consists of texts collected from the Wikipedia database in both French and English versions [17]. From these corpora, we generated tow vocabulary and language model configurations for each language.

The first configuration is a configuration without lexicon, which models the character sequences using a n-gram character model. The n-gram models are estimated by using Kneser-Ney smoothing of the MIT language model toolkit [20]. The second configuration is a word n-gram language model. Different dictionary size are used for training French n-grams language models that are trained on RIMES and French Wikipedia data sets. Similarly, several size of English dictionaries are used for training the English n-grams models on IAM and English Wikipedia data sets. The third configuration is a syllable n-gram model. The vocabularies are vocabularies of syllables obtained through the syllabification method proposed above, and the language models are n-grams of syllables estimated on the same corpora RIMES/IAM and Wikipedia (French/English).

*D. Recognition step*

Our recognition system is characterized by a tow pass decoding. The first pass processes the test sample by performing decoding according to the Viterbi algorithm with pruning over time (time synchronous Viterbi beam search). The optical models are used for the character model, or they are concatenated to form words or syllables of the lexicon to be used, depending on the recognition scenario.

According to the scenario, the decoding algorithm uses character, syllable or word bi-gram, to produce a network of character, syllable or word sequence hypotheses. Two important parameters guide this first decoding pass: they are the language model scaling parameter γ and the word insertion penalty parameter β that controls the insertion of too frequent short words. These two parameters need to be optimized for optimum coupling of the optical model with the considered language model, because these two models are estimated independently from each other during the training. The second decoding pass analyses the hypotheses network provided by the first pass using a language model of higher order n-gram which allows re-weighting the first hypotheses. This last step provides the final output recognition of the text line. When decoding we seek the $\widehat{W}$ word sequence that maximizes the posterior probability $P(W|S)$ among all possible sentences $W$. Using Bayes' formula and introducing the two hyper-parameters defined above, we finally arrive to the formula given in equation 2 which governs the decoding step. In this formula, $S$ represents the sequence of observations extracted from the image and $P(S|W)$ represents the likelihood that the characteristics $S$ are generated by the sentence $W$, it is deduced from the optical model. $P(W)$ is the prior probability of the sentence $W$, it is deduced from the language model.

$$\widehat{W} = argmax_w P(S|W) P(W)^\gamma \beta^{Length(W)} \qquad (2)$$

V. EVALUATION

To optimize and test the performance of our system, we used the RIMES validation data set that contains 764 rows retrieved from 100 images of paragraphs. Half of this validation set was used for optimizing the decoding parameters and the other half was used to evaluate the performance of the different configurations of the system. Furthermore, we used 1033 lines of IAM data set for evaluating the performance of our system at different configurations of the system.

We defined a first configuration of our system by combining the resources of the RIMES/IAM and Wikipedia data sets (French/English lexicons and language models) by adding RIMES/IAM word lexicons to its corresponding Wikipedia 10k, 20k and 40k most frequent word lexicons. This will allow to evaluate the three competing models (characters, syllables and words) under the condition of closed lexicon and language model at different size of lexicon. For each size of lexicon, a language model is specifically trained on Wikipedia and RIMES/IAM corpuses. The second configuration of our system evaluates the performance of different models in situations where the lexicon of the test data is partially covered by the system lexicon. In this case we will only use Wikipedia resources to determine the lexicons and their associated language models. We have retained the same Wikipedia lexicons size used in the closed lexicons mode (10K, 20K and 40K most frequent Wikipedia words).

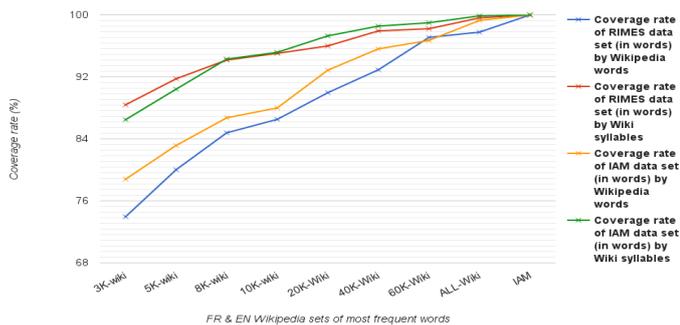

Fig. 1. Coverage rate of the RIMES and IAM data sets by the words and syllables lexicons of Wikipedia

Figure 1 provides the coverage rates (proportion of words present) of lexicon and RIMES / IAM databases for different lexicons of words and syllables extracted from the French / English Wikipedia corpus. We see that the data sets RIMES and IAM have an interesting out of vocabulary words rate when working with small lexicons size of Wikipedia vocabularies. The Figure 2 shows the lexicon size of syllables derived from words lexicons. We immediately observe the reduction of lexicons sizes (around two of three) by working with a syllabic model.

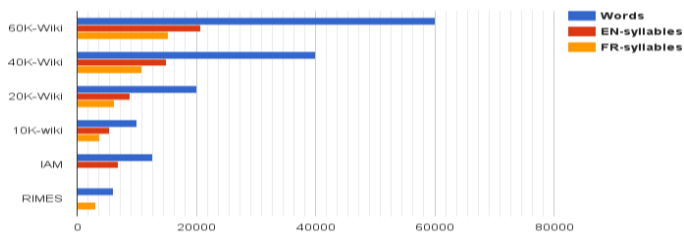

Fig. 2. Size of syllables lexicons derived form Wikipedia words lexicons

The word error rate (WER) is used to evaluate the performance of our recognition systems. Figure 3 illustrates the behavior of the system for each model (characters, syllables and words) and for the lexicon configurations that cover totally or partially the RIMES dataset. The behavior of the system for each model (character, syllables and words) is illustrated in Figure 4 for the lexicon configurations that cover totally or partially the IAM data set. In those figures we show the WER for the lexicons with decreasing coverage rates. On the left, we give the results of four lexicons including RIMES / IAM data set lexicon that necessarily have 100% word coverage rate on the test database, then we give the results for four lexicons that are build from Wikipedia only. The decreasing size of the French or English Wikipedia lexicons leads to a decreased coverage rate on RIMES data set (respectively from 97.94%, 96%, 95.04% for the syllabic model and 92.92% 89.96% 86.52% for the word model). The same effect has been observed regarding the Wikipedia lexicon size against the coverage rate on IAM data set (respectively from 98.56%, 97.3%, 95.17 for the syllabic model and 95.62%, 92.87%, 88% for the word model). In figure 3, we observe for the most closed lexicon configuration to the RIMES data set (first pointed result to the left), that the syllabic model gets the best performance (20.6% WER) which is slightly close to the word model (19.6% WER) and significantly better than the characters model performance (27.8% WER).

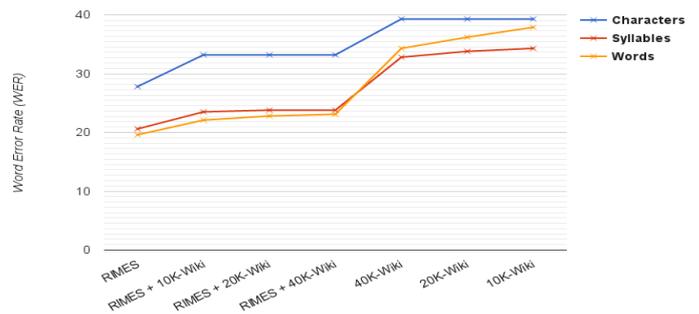

Fig. 3. WER (%) of the three models for different sizes of French lexicons

This trend is confirmed when increasing the vocabulary size and keeping a 100% coverage rate. We even observe that the syllabic model performance degrades less when the size of the lexicon increases, compared to the word model. These performances are obtained by using n-gram models of order 6 for the characters, 6 for the syllables, and 3 for the words. We can therefore conclude that for closed vocabulary, the syllabic model is a very good alternative to a lexical model, since it gets similar performance with a reduced complexity (reduced lexicon leads to reduced language model accordingly). In the other hand, we get the same observations from figure 4 which shows the test results on the English IAM data set. The most closed lexicon on IAM data set (the first pointed results at the left) shows that the syllabic model gets a performance (25.9% WER) close to the word model performance (24.5% WER). This behavior is also confirmed when increasing the lexicon size and keeping a 100% coverage rate, but contrary to the behavior of the French syllabic model, we observe a degradation as a function of the lexicon size. This can be explained by the increase of the number of syllables as shown in figure 4.

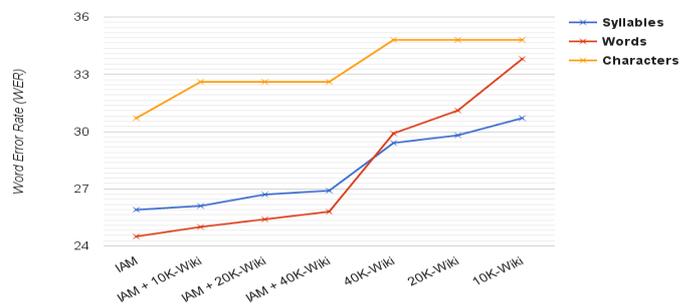

Fig. 4. WER (%) of the three models for different sizes of English lexicons

In Figures 3 and 4, the three tests using open vocabulary (Wikipedia lexicons only) show that syllabic model gets equal or superior performance to the word model. The models in figure 3 are equivalent when working with a lexicon of 40K (in this case the coverage rates are 92.92% for the word model, and 97.94% for the syllabic model respectively), whereas the models in figure 4 have almost superior performance with a lexicon of 40K (in this case the coverage rates are 95.62% for the word model, and 98.56% for the syllabic model respectively). In other configurations, the syllabic model gets better performance than the words model because it is able to cover OOV. Once again we see that the French syllabic model gets very stable performance regardless the lexicon size from which it was built. Thus, we can conclude, as we sought to demonstrate, that the syllabic model offers an interesting lexical coverage capacity, including out of vocabulary words while having a lower complexity than the lexical model.

Comparing the performance of our system with those reported in [2], we can notice that our syllabic recognition system (LITIS) is ranked in between the high rank state of the art systems and the low ranked ones. Table 2. Resume this comparison.

| Data Sets | WER (%) | | | | | | |
|---|---|---|---|---|---|---|---|
| | *LITIS* | *RWTH* | *Graves* | *Toselli* | *Espana* | *A2IA* | *Telecom ParisTech* |
| IAM | 25.9 | 13.3 | 25.9 | 25.8 | 22.4 | --- | --- |
| RIMES | 20.6 | 13.7 | --- | --- | --- | 15.2 | 31.2 |

TABLE II.  SYLLABIC MODEL PERFORMANCE COMPARISON.

## VI. CONCLUSION AND PERSPECTIVES

In this study we proposed French and English syllabic models for handwriting recognition. These models offer many advantages over the characters model and the word model. The advantages of this model are twofold. On the one hand, it is of limited complexity, since it works with a reduced lexicon of syllables. It follows an n-gram model of syllables which is itself more compact, so better parameterized, and therefore easier to optimize. On the other hand it offers superior performance than a lexical model when working with out of vocabulary words. To generate the syllabic model we stand on the lexique3 data set that propose an orthographic syllable modeling for the French language, and on the free English hyphenation dictionary, which propose the orthographic modelling of syllables for the English language. Other models could be evaluated for comparison. The question of the search for an optimal method for decomposing words into parts could also be explored. Moreover, it should be interesting to know if these models can offer the same interest on other languages. The interest in Arabic language has already been demonstrated as shown in the literature review. The door is open also for exploring the ability of constructing multi-lingual model based on the decomposition of different languages into syllables.

ACKNOWLEDGEMENT

Our thanks are to Ms. Elise Ryst and Mr. Christopher Coupeur who offered us their advices as language specialists.